%% file: main.tex
\documentclass[10pt,twocolumn,letterpaper]{article}

\usepackage[pagenumbers]{cvpr} 
\usepackage{csquotes}
\usepackage{colortbl}
\usepackage{wrapfig}
\usepackage{multirow}
\usepackage{arydshln}
\usepackage{amssymb}
\usepackage{pifont}
\usepackage{float} 
\usepackage{placeins} 

\usepackage{booktabs}
\usepackage{caption}

\usepackage{amsmath} 
\usepackage{tcolorbox} 
\usepackage{xcolor} 

\definecolor{mygreen}{rgb}{0,0.6,0} 

\newcommand{\xmark}{\ding{55}}%
\newcommand\blfootnote[1]{%
  \begingroup
  \renewcommand\thefootnote{}\footnote{#1}%
  \addtocounter{footnote}{-1}%
  \endgroup
}

\definecolor{cvprblue}{rgb}{0.21,0.49,0.74}
\usepackage[pagebackref,breaklinks,colorlinks,allcolors=cvprblue]{hyperref}

\def\paperID{15154} 
\def\confName{CVPR}
\def\confYear{2025}

\title{Empathetic Response in Audio-Visual Conversations \\ Using Emotion Preference Optimization and MambaCompressor}

\author{Yeonju Kim \quad\quad Se Jin Park \quad\quad Yong Man Ro$^\dagger$ \\
Integrated Vision and Language Lab, KAIST, South Korea \\
{\tt\small \{yeonju7.kim, jinny960812, ymro\}@kaist.ac.kr}
}

\begin{document}
\maketitle
\begin{abstract}

Chatbot research is advancing with the growing importance of chatbots in fields that require human interactions, such as customer support and mental health care. Despite these advancements, chatbots still face significant challenges in understanding subtle nuances and managing long conversation histories.  To address these issues, our study introduces a dual approach: firstly, we employ Emotional Preference Optimization (EPO) to train chatbots not only with correct responses but also with counter-emotional responses—those that are contextually similar but emotionally divergent. This training enables the model to discern fine nuance distinctions between correct and counter-emotional responses, thereby enhancing the quality of its responses. Secondly, we introduce MambaCompressor to effectively compress and manage extensive conversation histories, significantly reducing time and memory complexities while improving the chatbot's contextual understanding. Our comprehensive experiments across multiple datasets demonstrate that our model significantly outperforms existing models in generating empathetic responses and efficiently managing lengthy dialogues. 

\end{abstract}

\section{Introduction}
\blfootnote{$^\dagger$Corresponding author.}
\label{sec:intro}
A chatbot is a computer program that mimics conversation, creating text responses based on users' questions or inputs. Chatbots make it easy for users to get the services they need and have the potential to extend to traditionally human tasks such as counseling. These days chatbots are particularly useful in practical domains where human interaction is important, such as call center systems and mental health diagnostics. With the importance of chatbots, studies have enhanced human interaction by developing training methods that reflect people's preferences using reinforcement learning~\cite{ouyang2022training} and integrating audio-visual signals into the chatbot~\cite{alnuhait2023facechat,fei2024empathyear, chu2024towards} to understand the emotions and intentions of users' word. 

Despite significant advances in chatbot technology, two major challenges hinder the implementation of fully empathetic and context-aware systems. First, current models often struggle to capture subtle nuances of human communication, such as emotional undertones and complex intentions. These limitations prevent the model from imitating real conversations and perform poorly in situations where interactions beyond words must be understood. Second, managing long conversation history remains a challenging task. In the transformer-based language model, the computational complexity increases quadratically according to the input size~\cite{vaswani2017attention}. Consequently, other models often resort to truncation or summarization for long input sequences, which can lead to loss of information. In addition, long input sequences can disrupt the focus of the attention mechanism, hindering the generation of accurate answers. Likewise, transformer-based chatbots also have a problem of having to put a long conversation history as an input. 

As a solution to the first issue of failing to capture subtle nuances, we hypothesize that explicitly distinguishing between emotionally appropriate and inappropriate responses will help the model learn these fine nuance distinctions. To this end, we categorize responses as emotionally suitable or unsuitable and use these distinctions to train the model. While traditional chatbots are typically trained only on correct responses, our approach also includes emotionally inappropriate responses to promote a more nuance-sensitive and emotionally attuned output through Emotional Preference Optimization (EPO). EPO encourages the generation of emotionally aligned responses by training with both suitable and unsuitable examples. To implement this, we create our own set of counter-emotional responses, which are emotionally inappropriate and potential responses within opposite emotional contexts. We label emotions to construct emotionally contrasting situations and generate responses that reflect these opposite emotions, calling these \enquote{counter-emotional responses}. By using counter-emotional responses for EPO, the model learns a fine contrast between desirable and undesirable emotional tones, reinforcing its ability to discern and generate more contextually appropriate replies.

Additionally, we introduce a \enquote{MambaCompressor} to address the challenges of long sequence input in dialogue model. Inspired by Meteor~\cite{lee2024meteor}, we employ the MambaCompressor to summarize the conversation history, effectively managing the challenges associated with long input sequences while maintaining content.

In this study, we propose an empathetic and efficient chatbot that leverages advanced techniques. By utilizing Emotional Preference Optimization (EPO), the chatbot is capable of generating responses that are sensitive to nuances and empathetic. Unlike traditional methods, this model learns to discern subtle nuances through the differences between correct and counter-emotional responses. Additionally, we have developed an efficient model using the MambaCompressor summarization system. This allows the model to efficiently compress the conversation history while preserving the integrity of the content.

Our main contributions are summarized as follows:

\begin{itemize}
    \item This paper demonstrates a method that utilizes explicit emotion labels to generate counter-emotional responses, and by training the chatbot with both correct and counter-emotional responses through EPO, the chatbot can discern subtle nuances and accordingly generate appropriate responses.
    \item We introduce a MambaCompressor that summarizes the conversation history to handle the long input sequence issues. Through this, we can observe that using the MambaCompressor significantly reduces the input size by 81.6\%, memory usage by 3.5\%, and processing time by 28.5\% , while maintaining content.
    \item Extensive experiments on multiple datasets demonstrate that our model outperforms baselines in various metrics. Diverse analyses demonstrate that our model can understand subtle nuances and human emotions, enabling it to respond appropriately.
\end{itemize}

\section{Related Works}
\label{sec:related_works}

\subsection{Multi-modal Large Language Model}
With advancements in Large Language Models (LLMs), Multi-modal Large Language Models (MLLMs) have progressed significantly, allowing LLMs to process multiple modalities. Early research focused on image understanding, leading to architectures like BLIP-2~\cite{li2023blip}, LLaVA~\cite{liu2024visual}, and Qwen-VL~\cite{bai2023qwen}. This evolved into models like VideoLLaMA~\cite{zhang2023video}, LLaVA-NeXT~\cite{liu2024llava} and VideoChat~\cite{li2023videochat} for video comprehension, and SpeechGPT~\cite{zhang2023speechgpt} and Qwen-Audio~\cite{chu2023qwen, chu2024qwen2} for audio understanding. Some models, such as AnyGPT~\cite{zhan2024anygpt} and NextGPT~\cite{wu2023next}, have further advanced to integrate three or more modalities, handling diverse inputs and generating multi-modal outputs. Although research on MLLMs has been steadily emerging, chatbot research using multi-modality~\cite{alnuhait2023facechat,fei2024empathyear, chu2024towards} remains relatively underexplored. Our work focuses specifically on human interaction within video chatbot applications.

\subsection{Mamba}

Transformers~\cite{vaswani2017attention} are able to capture long-range dependencies through self-attention. However, they come with the problem of the quadratic scaling of computation with sequence length, which restricts the maximum input size. To address this limitation, state space model(SSMs) like S4~\cite{gu2021efficiently}, which maintains linear computational complexity while achieving competitive performance on long-range tasks have been proposed. Following S4, Mamba~\cite{gu2023mamba} and Mamba2~\cite{dao2024transformers} implement selective state space mechanisms to provide adaptive, dynamic state transitions to address the limitation of fixed dynamics in the previous SSMs. 

Beyond natural language processing, Mamba has demonstrated remarkable potential in various tasks. Vision Mamba~\cite{zhu2024vision} introduces a vision backbone with bidirectional Mamba to enhance image understanding, while VideoMamba~\cite{li2025videomamba, park2025videomamba} utilizes Mamba’s capabilities to model long-term dependencies in video. Additionally, Meteor~\cite{lee2024meteor} leverages Mamba to effectively compress lengthy input sequences into compressed features. Inspired by Meteor, we use MambaCompressor to condense conversation histories into compact embeddings, preserving essential information while ensuring computational efficiency.

\subsection{Preference Optimization}

Reinforcement Learning with Human Feedback (RLHF)~\cite{ouyang2022training} has been a prominent approach for aligning language models with human preferences. Typical RLHF training includes a supervised fine-tuning stage, followed by reward model training on preferred data and reinforcement learning optimization. This process comes with notable drawbacks, including the need to train multiple models and the computational cost of repeated sampling from the language model.

To address these limitations, Direct Preference Optimization (DPO)~\cite{rafailov2024direct} was introduced. DPO enables preference-based optimization directly on the language model without a separate reward model, simplifying the training process and reducing computational demands. This method has inspired the development of further preference optimization techniques~\cite{xu2024contrastive, zeng2024token, wu2024self}. One of them, SimPO~\cite{meng2024simpo} eliminates the need for a reference model and applying length normalization to reduce the tendency to generate longer but lower-quality sequences. Motivated by SimPO, we guide the language model to generate emotionally appropriate responses while avoiding emotionally inappropriate ones, thereby enhancing response quality in a more nuanced and emotionally aware manner.

\begin{figure*}[t]
	\begin{minipage}[b]{\linewidth}
		\centering		
        \centerline{\includegraphics[width=17cm]{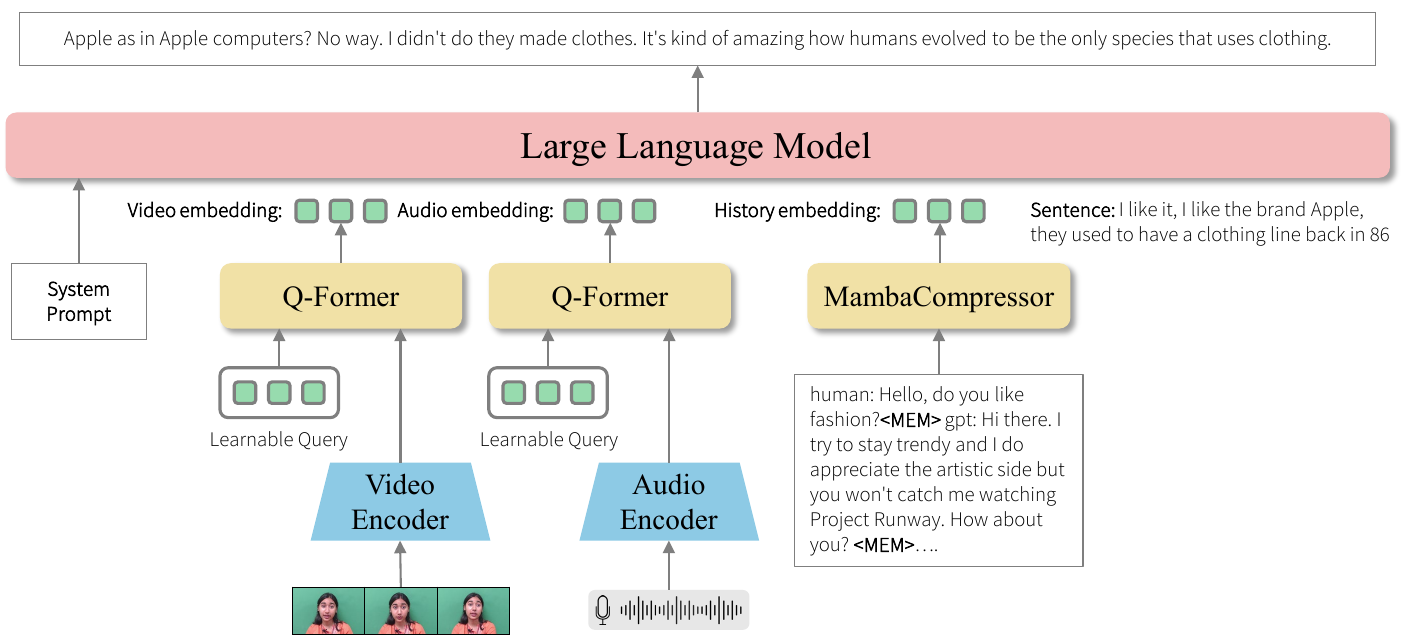}}
	\end{minipage}
        \vspace{-0.5cm}
	\caption{Overview of the proposed method. The model comprises an LLM backbone, an audio-visual emotion extractor, and MambaCompressor. The audio-visual emotion extractor extracts emotional information from the audio and video, while MambaCompressor summarizes the conversation history. These processed inputs, along with text inputs, are then fed into the LLM, which generates the response.}
	\label{fig:overview}
  \vspace{-0.2cm}
\end{figure*}

\section{Method}
\label{sec:method}

\subsection{Audio-Visual Emotion Extractor}
\label{subsec_av}

\begin{figure}[t]
	\centering
	\includegraphics[width=0.9\columnwidth]{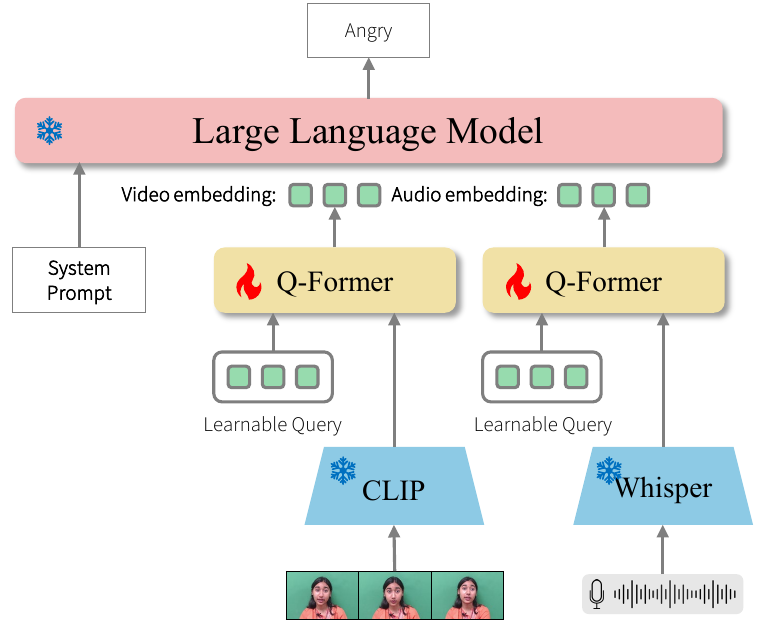}
	\caption{Audio-Visual Emotion Extractor. Audio-Visual Emotion Extractor is trained to extract emotion-related features from audio and video, with the goal of predicting the emotion category. We keep the LLM frozen and train the Q-former together with learnable queries.}
	\label{fig:av_train}
    \vspace{-0.4cm}
\end{figure}


It is challenging to discern a person's emotions from text alone, so we incorporate audio-visual information to capture emotional cues from the speaker. CLIP~\cite{radford2021learning} vision encoder is used to extract visual features from 8 randomly sampled frames of video, while Whisper~\cite{radford2023robust} is adopted to capture audio features. We extract hidden-state feature from the last layer of CLIP and Whisper. These features are then transformed by the Q-former in each encoder into emotion-specific embeddings. We train the Q-former and learnable queries on the emotion recognition task as illustrated in~\cref{fig:av_train}. With this task, the model aims to accurately recognize emotions by utilizing visual and auditory cues from the audio and video, such as facial expressions, gestures, and voice tone. During training, we employed three different input settings for the emotion recognition task: using only audio, only video, or both. This approach prevents reliance on any single modality, thereby enhancing the effectiveness of emotion recognition. 

Using the emotional video and audio embeddings, the LLM is guided by a system prompt to predict emotions across 7 categories: angry, disgust, fearful, happy, neutral, sad, and surprise. We utilize both audio and video as we have observed that relying on a single modality often proves insufficient for accurate emotion prediction as shown is~\cref{fig:emotion_recogntion}. When combined, the two modalities mutually compensate for each other in situations where one alone is insufficient.

\subsection{MambaCompressor}
\label{subsec_mamba}

\begin{figure}[t]
	\centering
	\includegraphics[width=0.75\columnwidth]{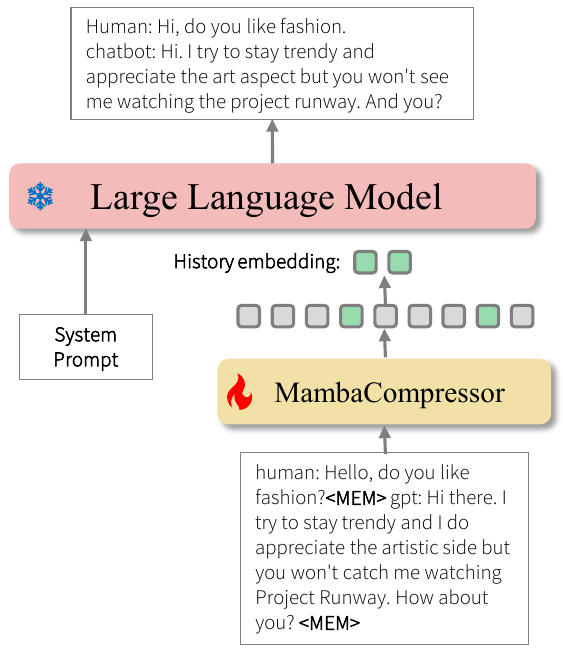}
	\caption{Illustration of the training process for the MambaCompressor. We trained the MambaCompressor with a frozen LLM on a conversation reconstruction task.    }
	\label{fig:m_train}
 \vspace{-0.6cm}
\end{figure}

Traditionally, models either processed truncated segments of conversation history, failing to capture the full context. Alternatively, they handled entire conversation histories, which significantly increased both time and memory complexities due to the lengthy input sequences. To address these issues, we have significantly reduced the size of the conversation history input by employing MambaCompressor. This reduction enhances efficiency from both time and memory perspectives, enabling the model to handle large-scale data more effectively without compromising performance. 

As depicted in~\cref{fig:m_train}, we insert $<$MEM$>$ tokens between utterances in the conversation, and each $<$MEM$>$ token is designed to encapsulate the content from the preceding utterance, essentially transforming a utterance into a compact output feature represented by the $<$MEM$>$ token. To achieve this, we train the MambaCompressor using a frozen LLM on the task of reconstructing the original utterances solely from the compacted $<$MEM$>$ token representations.

First, we train MambaCompressor to reconstruct a single utterance. Once the model has converged, we then finetune it to reconstruct multiple utterances. Through this progressive training process, MambaCompressor ultimately gains the ability to summarize lengthy conversations into few features. Therefore, the model can condense each utterance into a single feature, effectively mitigating the quadratic growth in computational complexity typically associated with processing lengthy conversation histories.

\subsection{Counter-Emotional Response Generation}
\label{subsec_counter_emotion}

\begin{figure}[t]
	\centering
	\includegraphics[width=\columnwidth]{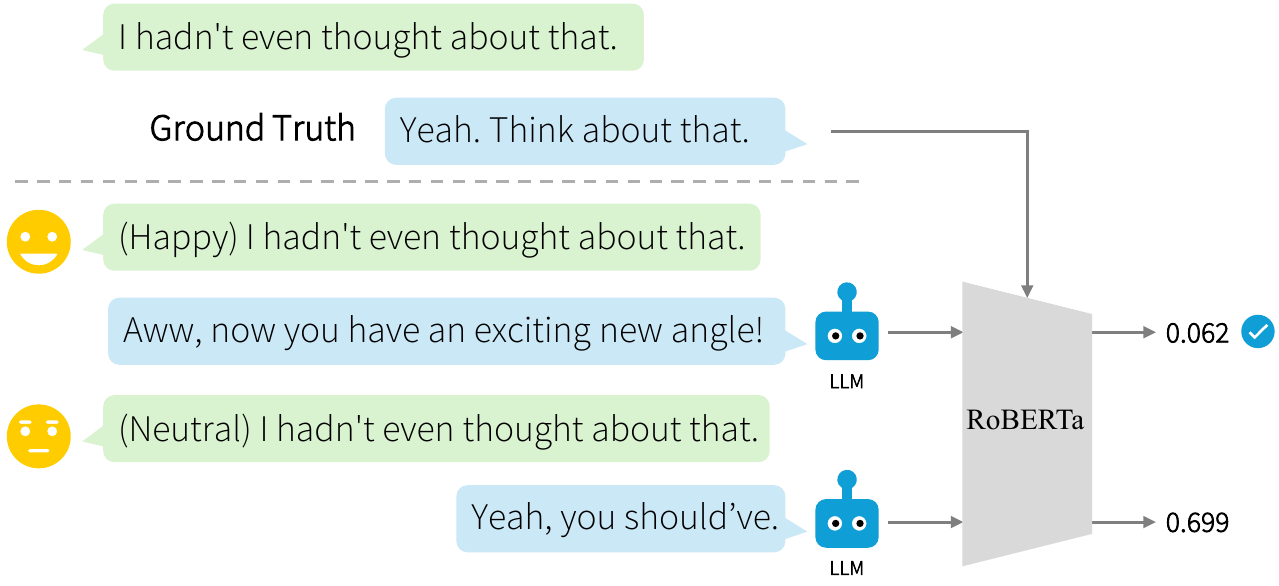}
	\caption{Counter-Emotional Response Generation. We categorize different emotional situations using brackets, and then generate responses based on these emotions. Subsequently, we calculate the emotional similarity between the ground truth and the generated response to select the counter-emotional response.}
 \vspace{-0.4cm}
	\label{fig:counter-emotional}
\end{figure}

To implement Emotional Preference Optimization (EPO), which will be detailed later in~\cref{subsec_epo}, it is essential to prepare counter-emotional responses. These are responses designed for situations with an opposite emotional context. We use the same LLM for generating counter-emotional responses as we do for training, because using a different model would result in responses that are too divergent, making them less effective for learning. 

As illustrated in~\cref{fig:counter-emotional}, we first label the emotional scenario of the given sentence, which assumes the emotion of the speaker uttering a given sentence. We then create seven distinct sentences, each labeled with one of the following emotions: angry, disgust, fearful, happy, neutral, sad, and surprise. These labeled sentences are then inputted into the LLM, which generates responses that are aware of the assigned emotional scenarios. We prompt the model to generate a response to the given sentence while considering the specific emotion. For example, as illustrated in~\cref{fig:counter-emotional}, in a situation where the statement is made in a happy tone, the response is more positive, such as \enquote{Aww, now you have an exciting new angle!} In contrast, in a Neutral situation, the response is more monotone, as in \enquote{Yes, you should've.} 

Among the generated responses, we select the most emotionally inappropriate sentence. To achieve this, we calculate the emotional similarity between the ground truth and the generated responses by computing the cosine similarity of embeddings from RoBERTa~\cite{liu2019roberta} trained on the GoEmotions dataset~\cite{demszky2020goemotions}. RoBERTa is the model known for its effectiveness in sentence classification, and when trained on the GoEmotions emotion classification dataset, it becomes proficient at identifying the emotional content of sentences. As a result, the features extracted by RoBERTa effectively capture the emotional information of the sentences. The response with the lowest similarity is selected as the counter-emotional response and is used in EPO, as detailed in~\cref{subsec_epo}.

\subsection{Emotional Preference Optimization(EPO)}
\label{subsec_epo}

Previously, models are trained with a single correct sentence, often failing to capture subtle nuances. To address this, we have trained our model using Emotional Preference Optimization (EPO). This involves learning from a correct sentence along with a counter-emotional sentence—similar in content but slightly different emotionally—allowing the LLM to discern these subtle nuances more effectively. Emotional Preference Optimization (EPO) treats the LLM itself as a reward model and operates without the need for a reference model, similar to SimPO~\cite{meng2024simpo}. The learning process of EPO involves increasing rewards for correct responses and decreasing rewards for counter-emotional responses. In this approach, EPO employs the average log likelihood as the reward metric, as shown in~\cref{eqn:reward_epo}. In~\cref{eqn:reward_epo}, $x$ is input sequence, $y$ is target sequence, $\pi_\theta$ is LLM model, and $\beta$ controls the reward scaling between correct and counter-emotional responses.

\begin{equation}
\label{eqn:reward_epo}
r_{EPO}(x,y)= \frac{\beta}{|y|} \sum_{i=1}^{|y|} \log \pi_\theta(y_i \mid x,y_{<i}).
\end{equation} 
By integrating this reward formulation into the Bradley-Terry (BT) ranking objective~\cite{bradley1952rank}, we can derive the loss function as shown in~\cref{eq:epo}. Using this objective, we train the LLM to maximize the reward for correct (appropriate) responses $y_a$, making it higher than that for counter-emotional (inappropriate) responses $y_i$. We incorporate a target reward margin term, $\gamma$, which boosts the rewards for appropriate responses by a margin of $\gamma$. $\mathcal{D}$ indicates that the dataset consists of correct and counter-emotional responses and $\sigma$ is sigmoid function.
\begin{multline}
\label{eq:epo}
\mathcal{L}_{EPO}(\pi_\theta) \\
= - \mathbb{E}_{(x, y_a, y_i) \sim \mathcal{D}}\left[ \log \sigma \left( r(x, y_a) - r(x, y_i) - \gamma \right) \right].
\end{multline}

We finetune the pretrained LLM using EPO loss in~\cref{eq:epo} and the standard autoregressive loss. This additional EPO training enables the LLM to discern subtle nuances between correct and counter-emotional responses, allowing it to respond more accurately and empathetically.

\input{tables/table1}

\section{Experiments}
\label{sec:experiments}
\subsection{Datasets}
We evaluate our method using benchmarks similar to real-world conversations with emotion labels.

\textbf{MultiDialog}~\cite{park2024let} is a large-scale audio-visual dataset containing approximately 370 hours of dialogue, with around 9,000 conversations exchanged between six speaker pairs. The scripts are derived from Topical-Chat~\cite{gopalakrishnan2023topical} and each utterance includes emotion annotations. We use the train split for training and the test-freq split for test.

\textbf{Multimodal Emotion Lines Dataset (MELD)}~\cite{poria2018meld} is a audio-visual conversation dataset based on conversations from the TV series Friends. This dataset contains approximately 1,433 conversations and 13,000 utterances, each annotated with emotion labels.

\textbf{IEMOCAP}~\cite{busso2008iemocap} is a audio-visual conversation dataset. The dataset consists of approximately 12 hours of audio-visual data involving 10 actors, who read emotional scripts or act out impromptu situations. Each utterance is annotated with emotion labels. This dataset does not come with predefined train and test splits, so we performed a random split, allocating 90\% of the data for training.

\subsection{Implementation Details}
We adopt a pre-trained Whisper tiny model~\cite{radford2023robust} for the audio encoder and pre-trained CLIP-ViT-B/32~\cite{radford2021learning} for video encoder. Q-formers are transformer encoders with 6 layers and 8 attention heads, utilizing 32 learnable queries to extract emotional features. When we train the model on an emotion recognition task to generate emotion-related features, we add RAVDESS~\cite{livingstone2018ryerson} and CREMA-D~\cite{cao2014crema} to the training data, because the data contains good facial expression and speech about the emotion. For MambaCompressor, we leverage the pretrained Mamba-370M~\cite{gu2023mamba} model and finetune it on our training dataset. We introduce a $<$MEM$>$ token to the vocabulary and insert a $<$MEM$>$ token at the end of every utterance, using its output feature to summarize the preceding content. 

For the large language model (LLM), we use Qwen2.5-7B-Instruct~\cite{yang2024qwen2}. During the fine-tuning of the LLM on the conversation dataset, we apply LoRA~\cite{hu2021lora} with a rank of 128 and an alpha of 256, while keeping the pre-trained audio-visual emotion extractor and MambaCompressor frozen. For training with EPO, we set $\beta$ to 2 and $\gamma$ to 0.5, and the learning rate to $1e^{-6}$. Before training the model with EPO, we first train our LLM on dialogue task, alongside the memory reconstruction task. This joint training prevents the model from losing the memory understanding capability acquired in~\cref{subsec_mamba}. 

\subsection{Baseline Methods}
We compare our model with current state-of-the-art models, including Qwen-2.5-Instruct~\cite{yang2024qwen2}, Claude 3.5 Haiku, Gemini 1.5 Flash~\cite{team2024gemini}, GPT-4o~\cite{achiam2023gpt}, Qwen-Audio-Chat~\cite{chu2023qwen}, and LLaVA-NeXT~\cite{liu2024llava}. We assess both the pretrained Qwen-2.5-Instruct and the fine-tuned version using training data. Claude 3.5 Haiku, Gemini 1.5 Flash, and GPT-4o are proprietary models known for their powerful performance due to large model sizes and extensive training. However, their weights are not publicly available, which prevents any further training. For a fair comparison, by providing the conversation history in a prompt, we enable these proprietary models that have not been directly trained on training data to perform in-context learning. Qwen-Audio-Chat process audio and text input and outputs text. LLaVA-NeXT takes audio and text as input and produces text as output. We fine-tune Qwen-Audio-Chat and LLaVA-NeXT on the training dataset using LoRA.

\subsection{Evaluation Metrics}
We use the following automatic evaluation metrics to assess our model: BertScore~\cite{zhang2019bertscore}, EmotionScore, and Dist-1~\cite{li2015diversity}. BERTScore is a metric for evaluating text generation quality that captures semantic similarity more effectively than traditional n-gram based methods like BLEU~\cite{papineni2002bleu} and ROUGE~\cite{lin2004rouge}, by computing cosine similarity between embeddings of generated and reference text using a pretrained BERT model. EmotionScore is a metric used to assess the emotional appropriateness of a generated sentence. Like BERTScore, it calculates the cosine similarity between the features of generated sentences and reference sentences using RoBERTa base model\footnote{https://huggingface.co/SamLowe/roberta-base-go\_emotions} trained on GoEmotions dataset. A higher EmotionScore indicates a high similarity with the reference sentence in terms of emotion. Dist-1 measures the diversity of text generation by calculating the ratio of unique 1-grams in the output. A higher Dist-1 score indicates less duplication and more varied vocabulary. Metrics are selected to assess whether the response is meaningfully appropriate using BertScore, emotionally congruent using EmotionScore, and sufficiently diverse with Dist-1.

\section{Results and Analysis}

\subsection{Automatic Evaluation of Dialogue Generation}

The results in~\cref{table1} show the effects of our model on generating responses in terms of semantic appropriateness, emotional appropriateness, and diversity. Our model demonstrates robust performance across different datasets, indicating its effectiveness in various conversational contexts.

The BertScore results, which are the highest across all benchmarks except for MELD, suggest that our model is good at understanding the flow of conversation and generating semantically appropriate responses. This indicates that MambaCompressor condenses input utterances into a few features while still retaining the context. Furthermore, by enabling the model to fully grasp and respond to the historical context more effectively, it performs better than the comparison model. A notable improvement in the IEMOCAP dataset, which consists of spontaneous dialogue from video interactions and closely mimics actual dialogue situations, suggests our model is well-suited for real-world conversational scenarios. 

The EmotionScore results indicate that our model excels at generating emotionally appropriate responses across all benchmarks. This performance suggests that our approach enables the model to discern subtle emotional nuances from audio-visual signals and text, and respond accordingly.

The high Dist-1 scores across benchmarks demonstrate our model's ability to generate diverse responses, which is essential for natural conversation. It shows that, unlike previous models which tended to provide uniform responses, our model can respond more diversely by adapting to different emotional situations, leveraging subtle emotional nuances from audio-visual signals and text.

\begin{figure}[t]
	\centering
	\includegraphics[width=\columnwidth]{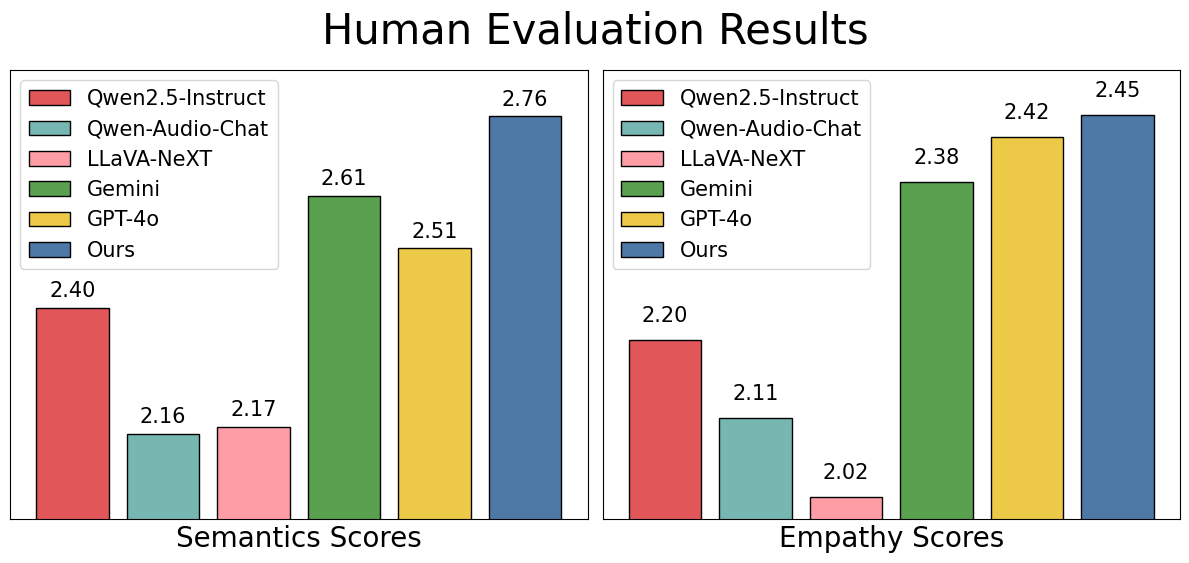}
        \vspace{-0.7cm}
	\caption{Human Evaluation. Results of the human evaluation comparing different models on two metrics: Semantics Scores and Empathy Scores. Higher scores indicate better performance.}
    \vspace{-0.3cm}
	\label{fig:human_eval}
\end{figure}

\subsection{Human Evaluation}

The human evaluation was conducted on the Prolific platform, with participants residing in English-speaking countries. In the experiment, participants were provided with the entire video containing the conversation history and were asked to rate the responses generated by the models using a three-point scale: good, medium, and bad. Scores were assigned as follows: good = 3, medium = 2, and bad = 1, and the average was calculated based on these values. Participants were asked two questions: \enquote{How semantically appropriate is the sentence as a continuation of the conversation?} and \enquote{How well does this sentence evoke empathy with the final statement?} The results are presented in~\cref{fig:human_eval} as empathy scores and semantic scores.

The results showed that our approach outperformed other methods in terms of contextual appropriateness, and emotional appropriateness. This finding aligns with the results of automatic evaluation in~\cref{table1}, demonstrating the superiority of our approach not only in metrics but also in practical effectiveness.

\subsection{Ablation study}
\input{tables/table2}

The results in~\cref{table2} display the ablation study conducted on our method using the IEMOCAP benchmark. We tested five model versions, all based on Qwen2.5-7B-Instruct model. The first model is not finetuned, the second is finetuned, the third includes MambaCompressor, the fourth incorporates an audio-visual emotion extractor, and the last version is a model trained with EPO applied. When the MambaCompressor is not used, we provide conversation history as a prompt to aid contextual understanding.

The results demonstrate the effectiveness of our proposed ideas. Comparing the second and third rows, we see an increase in BertScore, suggesting that even with input size compressed by the MambaCompressor, the model can maintain context. However, the EmotionScore decreases slightly in the same comparison, indicating that while context is preserved, some subtle emotional nuances may be missed. When comparing the third and fourth rows, the EmotionScore improves, suggesting that using audio-visual signals enables the model to generate more emotionally resonant responses. Finally, in the last row, we observe that the use of EPO results in increased BertScore, EmotionScore, and Dist-1, highlighting the enhanced overall performance.

\subsection{Emotion Recogntion Accuracy}

\begin{figure}[t]
	\centering
	\includegraphics[width=0.7\columnwidth]{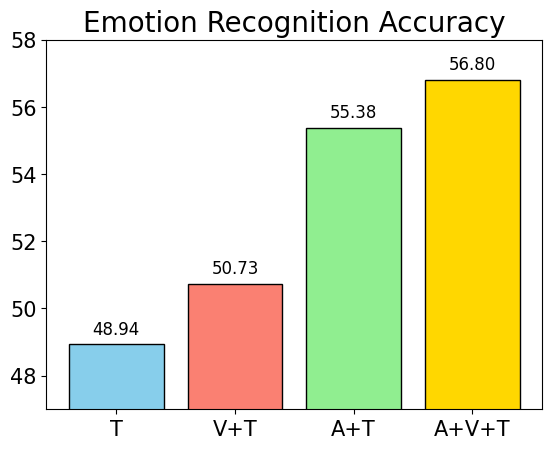}
        \vspace{-0.2cm}
	\caption{Emotion Recognition Accuracy: This figure shows the accuracy (\%) for emotion recognition using different input modalities. \enquote{T} represents text input, \enquote{V} represents video input, and \enquote{A} represents audio input. Combinations of inputs (e.g., A+T, A+V+T) demonstrate the impact of multimodal inputs on accuracy, with A+V+T yielding the highest performance.}
	\label{fig:emotion_recogntion}
\end{figure}

\cref{fig:emotion_recogntion} shows the result of the experiment designed to verify if adding audio and video features helps in identifying the emotion of a sentence. We utilized the audio encoder and video encoder trained as described in~\cref{subsec_av} to extract audio and video features for this purpose. The results indicate that using both audio and video features improves accuracy by 7.86\% compared to using text alone. It is also observed that using both audio and video together is more effective for emotion recognition than using only single video or single audio. This demonstrates that subtle emotional nuances, which may not be conveyed through text, can be captured effectively through audio and video inputs.

\subsection{Efficiency Advantage of MambaCompressor}
\input{tables/table3}

\cref{table3} shows the results of an experiment conducted using only samples with long conversation histories of 20 or more turns from the MultiDialog dataset. We reported the average input size for prompts, as well as the memory and time required to process the prompts. To ensure a fair comparison, we measured the resources used solely for prompt processing by configuring the setup to avoid generating responses during this stage. To evaluate the quality of generated responses, we obtained responses using a separate setup configured for response generation, and evaluated them using BertScore for quality assessment.

The results, shown in~\cref{table3}, demonstrate that using MambaCompressor enables efficient resource utilization, particularly for handling long contexts. Specifically, input tokens were reduced by 81.6\%, memory usage decreased by 3.50\%, and time consumption dropped by 28.5\%. We also observe that while BertScore is even slightly improved, resource usage was significantly reduced. These findings demonstrate that MambaCompressor can substantially reduce resource usage while maintaining response quality, especially when the conversation history is lengthy.

\subsection{Case studies}

\begin{figure}[t]
    \centering
    \begin{subfigure}[b]{\columnwidth}
        \centering
        \includegraphics[width=\columnwidth]{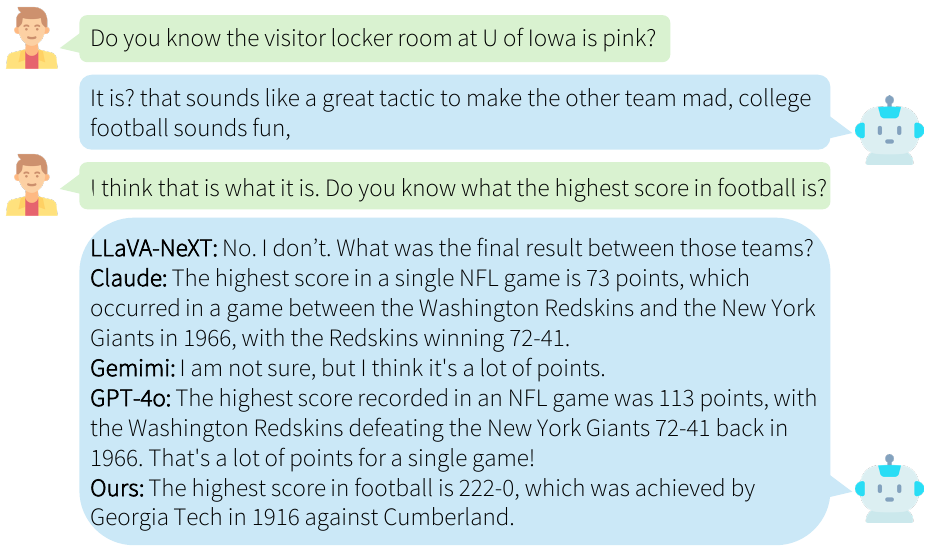}  
        \caption{}
        \label{fig:case1}
    \end{subfigure}
    \begin{subfigure}[b]{\columnwidth}
        \centering
        \includegraphics[width=\columnwidth]{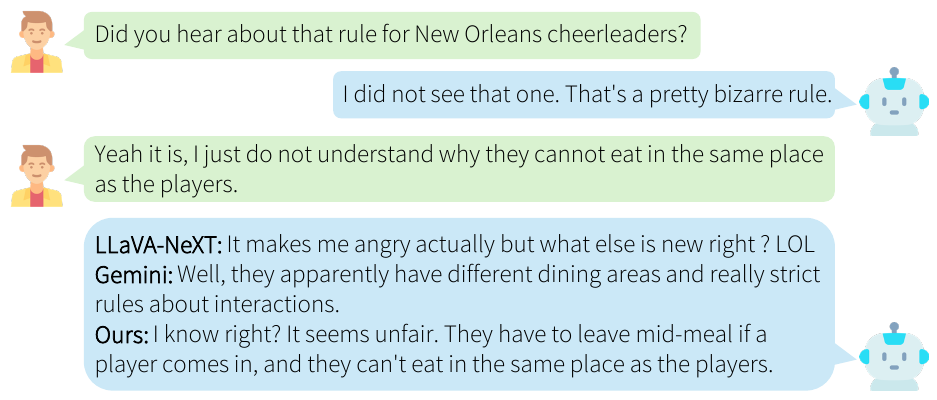}  
        \caption{}
        \label{fig:case2}
    \end{subfigure}
    \begin{subfigure}[b]{\columnwidth}
        \centering
        \includegraphics[width=0.95\columnwidth]{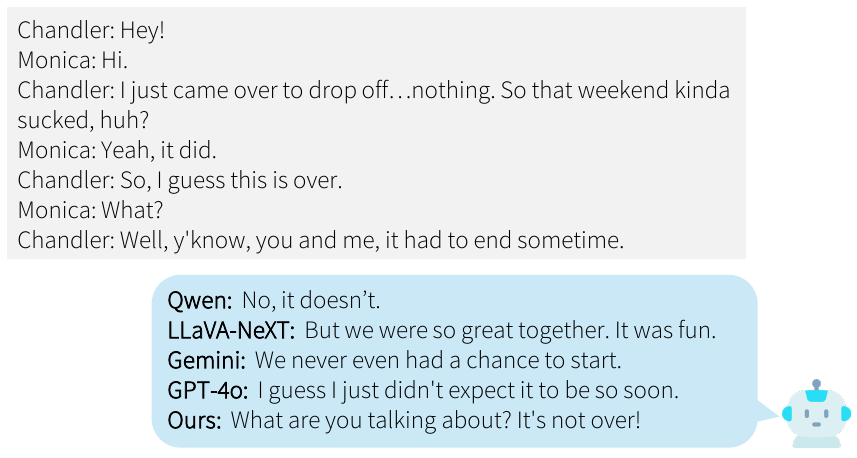}  
        \caption{}
        \label{fig:case3}
    \end{subfigure}    
        \begin{subfigure}[b]{\columnwidth}
        \centering
        \includegraphics[width=0.95\columnwidth]{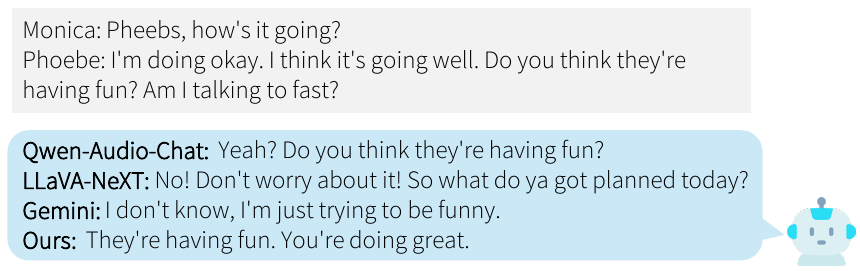}  
        \caption{}
        \label{fig:case5}
    \end{subfigure}   
    \caption{Case studies showing how our model excels in understanding emotions and noticing the fine details in conversations.}
    \vspace{-0.2cm}
    \label{fig:case_studies}
\end{figure}

We show our generation results in~\cref{fig:case_studies}. In~\cref{fig:case1}, the conversation begins with a discussion about the Iowa football team and then smoothly transitions into a broader conversation about college football teams. Therefore, the user's final question, \enquote{Do you know what the highest score in football is?} can be interpreted as asking about the highest score across all football leagues, not just the NFL. LLaVA-NeXT responds with \enquote{I don't know,} while Claude, Gemini, and GPT-4o focus on the highest score in the NFL. However, our model remembers the context about college football teams and mentions the 222-0 game between Georgia Tech and Cumberland in the college football league as the highest score. In~\cref{fig:case2}, there is a discussion about a strange rule preventing cheerleaders from sitting and eating with the players. The user expresses frustration, indicating discomfort through the video. In this context, LLaVA-NeXT downplays the issue by saying it’s annoying but not surprising, showing a lack of alignment with the conversation flow and the user's emotions. Gemini simply reconfirms the factual information without showing empathy. However, our model responds emotionally, describing the rule as unfair, showing empathy and aligning with the user's emotions. In~\cref{fig:case3}, Chandler and Monica exchange dialogue where Chandler states that their relationship is over. Monica responds with a surprised "What?", indicating shock at her unexpected words. In the final Chandler's word, Chandler expresses sadness in the video about the relationship falling apart. Here, Qwen and LLaVA-NeXT respond rather neutrally, while Gemini reacts with overly sarcastic comments toward Chandler's sadness. GPT-4o captures Monica’s surprise but responds in an emotionally flat manner. In contrast, our model replys "What are you talking about? It's not over," continues with Monica's shocked reaction and delivering a more emotionally appropriate response. In~\cref{fig:case5}, Phoebe shows concern about the people and appears nervous in the video. Qwen-Audio-Chat and Gemini seem to respond in a literal manner, understanding her words without recognizing the underlying anxiety. LLaVA-NeXT understands that Phoebe is worried but does not offer any supportive response. In contrast, our model reassures her by saying she's doing well, which demonstrates an understanding of her nervousness and having empathy. These examples demonstrate that our model understands the overall content of conversations well and responds with empathy.

\section{Conclusion}
\label{sec:conclusion}

This paper addresses the issues of traditional chatbots by integrating Emotional Preference Optimization (EPO) and MambaCompressor. Traditional chatbots have struggled with understanding subtle emotional nuances and managing long conversation histories. By training chatbot model with both correct and counter-emotional sentences through EPO, our approach enables the chatbot to respond more sensitively to nuances and emotions. Furthermore, MambaCompressor significantly reduces input size, lowering time and memory complexities while also enabling the chatbot to understand conversation history more accurately by compressing lengthy histories. As a result, our chatbot demonstrates a profound capability to comprehend overall conversation content and generate more empathetic responses. 

{
    \small
    \bibliographystyle{ieeenat_fullname}
    \bibliography{main}
}

\input{sec/X_suppl}

\end{document}

%% file: tables/table1.tex
\begin{table*}
\centering
\caption{The result of Automatic Evaluation of Dialogue Generation. An upward arrow indicates that a higher value is better, while a downward arrow signifies that a lower value is preferable. \enquote{$\dagger$} represents models that were finetuned on the training dataset.}
\vspace{-0.2cm}
\resizebox{\linewidth}{!}{
\begin{tabular}{lccccccccc}
\hline
\multirow{2}{*}{Model} & \multicolumn{3}{c}{MultiDialog} & \multicolumn{3}{c}{MELD} & \multicolumn{3}{c}{IEMOCAP} \\ \cmidrule(lr){2-4} \cmidrule(lr){5-7} \cmidrule(lr){8-10}
& BertScore$\uparrow$ & Emotion$\uparrow$ & Dist-1$\uparrow$ & BertScore$\uparrow$ & Emotion$\uparrow$ & Dist-1$\uparrow$ & BertScore$\uparrow$ & Emotion$\uparrow$ & Dist-1$\uparrow$ \\ \hline
Qwen2.5-7B-Instruct & 0.732 & 0.398 & 0.821 & 0.835 & 0.426 & 0.933 & 0.844 & 0.422 & 0.949 \\  
Qwen2.5-7B-Instruct$^\dagger$ & 0.837 & 0.404 & 0.931 & 0.866 & 0.447 & 0.953 & 0.864 & 0.503 & 0.973 \\
Qwen-Audio-Chat$^\dagger$ & 0.844 & 0.354 & 0.953 & 0.853 & 0.458 & 0.978 & 0.865 & 0.486 & 0.943 \\
LLaVA-NeXT$^\dagger$ & 0.845 & 0.374 & 0.960 & 0.853 & 0.453 & 0.980 & 0.840 & 0.424 & 0.975  \\
Claude 3.5 haiku & 0.848 & 0.401 & 0.946 & 0.862 & 0.417 & 0.979 & 0.861 & 0.391 & 0.962 \\
Gemini 1.5 flash & 0.852 & 0.396 & 0.958 & 0.833 & 0.473 & 0.947 & 0.864 & 0.442 & 0.940 \\
GPT-4o & 0.855 & 0.410 & 0.931 & \textbf{0.871} & 0.411 & 0.980 & 0.868 & 0.369 & 0.950 \\
\cellcolor[HTML]{F0F0F0}Ours$^\dagger$ & \cellcolor[HTML]{F0F0F0}\textbf{0.859} & \cellcolor[HTML]{F0F0F0}\textbf{0.459} & \cellcolor[HTML]{F0F0F0}\textbf{0.964} & \cellcolor[HTML]{F0F0F0}0.870 & \cellcolor[HTML]{F0F0F0}\textbf{0.485} & \cellcolor[HTML]{F0F0F0}\textbf{0.982} & \cellcolor[HTML]{F0F0F0}\textbf{0.907} & \cellcolor[HTML]{F0F0F0}\textbf{0.593} & \cellcolor[HTML]{F0F0F0}\textbf{0.983} \\ \hline
\end{tabular}
\label{table1}
}
\end{table*}

%% file: tables/table2.tex
\begin{table}
\caption{Ablation study on finetuning (FT), MambaCompressor (MC), audio-visual emotion extractor (AV), and EPO on the IEMOCAP benchmark.}
\vspace{-0.2cm}
\centering
\resizebox{0.9\linewidth}{!}{
\begin{tabular}{cccccccc}
\hline
FT & MC & AV & EPO & BertScore$\uparrow$ & Emotion$\uparrow$ & Dist-1$\uparrow$ \\ \hline
\xmark & \xmark & \xmark & \xmark &  0.844 & 0.422 & 0.949  \\
\checkmark & \xmark & \xmark & \xmark & 0.864 & 0.503 & 0.973 \\
\checkmark & \checkmark & \xmark & \xmark &  0.881 & 0.484 & 0.961 \\
\checkmark & \checkmark & \checkmark & \xmark &  0.882 & 0.525 & 0.963 \\
\cellcolor[HTML]{F0F0F0}\checkmark & \cellcolor[HTML]{F0F0F0}\checkmark & \cellcolor[HTML]{F0F0F0}\checkmark & \cellcolor[HTML]{F0F0F0}\checkmark & \cellcolor[HTML]{F0F0F0}\textbf{0.907} & \cellcolor[HTML]{F0F0F0}\textbf{0.593} & \cellcolor[HTML]{F0F0F0}\textbf{0.983} \\ \hline
\end{tabular}
\label{table2}
}
\end{table}

%% file: tables/table3.tex
\begin{table}
\caption{Memory and Time Efficiency Improvements with MambaCompressor(MC). Time is the average per-sample duration.}
\vspace{-0.2cm}
\centering
\resizebox{1.0\linewidth}{!}{
\begin{tabular}{clllc}

\hline
MC & Input Size & Memory(GB) & Time(sec) & BertScore \\ \hline
\xmark & 606.23 & 16.30 & 0.1112 & 0.844 \\
\cellcolor[HTML]{F0F0F0}\checkmark & \cellcolor[HTML]{F0F0F0}111.52{\color[HTML]{009901}(-81.6\%)} & \cellcolor[HTML]{F0F0F0}15.73{\color[HTML]{009901}(-3.50\%)} & \cellcolor[HTML]{F0F0F0}0.0795{\color[HTML]{009901}(-28.5\%)} & \cellcolor[HTML]{F0F0F0}\textbf{0.848} \\ \hline
\end{tabular}
\label{table3}
}
\end{table}

%% file: sec/X_suppl.tex
\clearpage
\setcounter{page}{1}
\maketitlesupplementary

\section{Comprehensive Training Instructions}
\label{sec:instruction}

During the training process, we guided the model to perform tasks at each stage using instructions. This section presents the instructions used during training.

\subsection{Audio-Visual Emotion Extractor Training Instruction}
\vspace{-0.5cm}
\begin{table}[H]  
\centering
\begin{minipage}{0.95\linewidth}  
\centering
\begin{tcolorbox}
\small
{\color[HTML]{3531FF} \textbf{Audio-Visual Emotion Extractor \\Training Instructions:}}\\
You are a language model trained to perform emotion recognition tasks. Your goal is to take speech embeddings and video embeddings as input and predict the corresponding emotions. Please answer in one word: happy, sad, surprised, calm, neutral, disgust, angry, and fearful.
\\
speech embeddings: $<$audio$>$
\\
video embeddings: $<$video$>$
\end{tcolorbox}
\vspace{-0.4cm}
\caption{Training instuction of audio-visual emotion extractor.}
\vspace{-0.3cm}
\label{table:prompt_av}
\end{minipage}
\end{table}
When training the audio-visual emotion extractor, the model receives audio and video as input and performs a task to classify emotions into one of seven categories. To facilitate this, we provided a task description along with answer choices as shown in~\cref{table:prompt_av}. The placeholders $<$audio$>$ and $<$video$>$ were replaced with audio embeddings and video embeddings, respectively. 

\subsection{MambaCompresor Training Instruction}
\vspace{-0.3cm}
\begin{table}[H]  
\centering
\begin{minipage}{0.95\linewidth}  
\centering
\begin{tcolorbox}
\small
{\color[HTML]{3531FF} \textbf{MambaCompressor Pretraining Instructions:}}\\
You are a language model trained to reconstruct the sentence. Your goal is to take memory embeddings as input and reconstruct the original sentence. Please answer with only the reconstructed sentence without any other explanation.
\\
memory embeddings: $<$memory$>$
\\
\\
{\color[HTML]{3531FF} \textbf{MambaCompressor Finetraining Instructions:}}\\
You are a language model trained to reconstruct the utterances. Your goal is to take memory embeddings as input and reconstruct the original utterances. One memory embedding represents one utterance. Please answer with only the reconstructed utterances without any other explanation.
\\
memory embeddings: $<$memory$>$
\end{tcolorbox}
\vspace{-0.4cm}
\caption{Training instruction of MambaCompressor.}
\vspace{-0.3cm}
\label{table:prompt_mem_1}
\end{minipage}
\end{table}
In~\cref{table:prompt_mem_1}, the MambaCompressor initially performed a pretraining task focused on reconstructing a single utterance. It was later finetuned to reconstruct multiple utterances as~\cref{table:prompt_mem_1}. This approach follows a coarse-to-fine strategy, progressively enhancing the model's capabilities. In this context, the $<$memory$>$ placeholder is replaced with MambaCompressor's memory embeddings.

\subsection{Instruction for Audio-Visual Dialogue Task with MambaCompressor}
\begin{table}[H]  
\centering
\begin{minipage}{0.95\linewidth}  
\centering
\begin{tcolorbox}
\small
{\color[HTML]{3531FF} \textbf{Instruction for Audio-Visual Dialogue Task with MambaCompressor:}}\\
You are a language model trained to generate responses that are relevant to the input sentence. Your responses take into account the context from previous dialogues, as well as emotional cues captured through speech and video embeddings. The history includes information from past interactions. The speech and video embeddings provide emotional cues in the tone, accent, speech speed, facial expressions, and gestures of the input sentence. Please just generate a response without any explanation.
\\
history: $<$memory$>$
\\
speech embeddings: $<$audio$>$
\\
video embeddings: $<$video$>$
\\
sentence: $<$sentence$>$
\end{tcolorbox}
\vspace{-0.4cm}
\caption{Training instruction for the audio-visual dialogue task with MambaCompressor.}
\label{table:prompt_dialog}
\end{minipage}
\end{table}

To fine-tune the LLM for dialogue tasks using Audio-Visual information and MambaCompressor embeddings, we trained it with instruction as shown in~\cref{table:prompt_dialog}. For this task, the LLM received MambaCompressor embeddings, audio embeddings, video embeddings, and a text sentence as input to generate the corresponding response.

\section{Counter-Emotional Response Details}
\label{sec:counter_emotion}

The counter-emotional response generation task focuses on creating responses that align with the specified emotional tone of a given sentence. As illustrated in~\cref{table:prompt_counter_emotional}, the model is instructed to generate responses by considering both the given sentence and the associated emotion label. For example, when the emotion label is \enquote{Happy}, the model generates a response for the sentence \enquote{Do you listen to radio at all?} in a cheerful and positive tone. To achieve this, we used Qwen2.5-7B-Instruct, the backbone of our model, to perform self-generation and obtain candidate counter-emotional responses. This process ensured that the generated responses avoid overly divergent outputs. The emotion labels used in this task include Happy, Sad, Surprised, Calm, Neutral, Disgust, Angry, and Fearful.
\begin{figure}[H]
    \centering
    \includegraphics[width=\columnwidth]{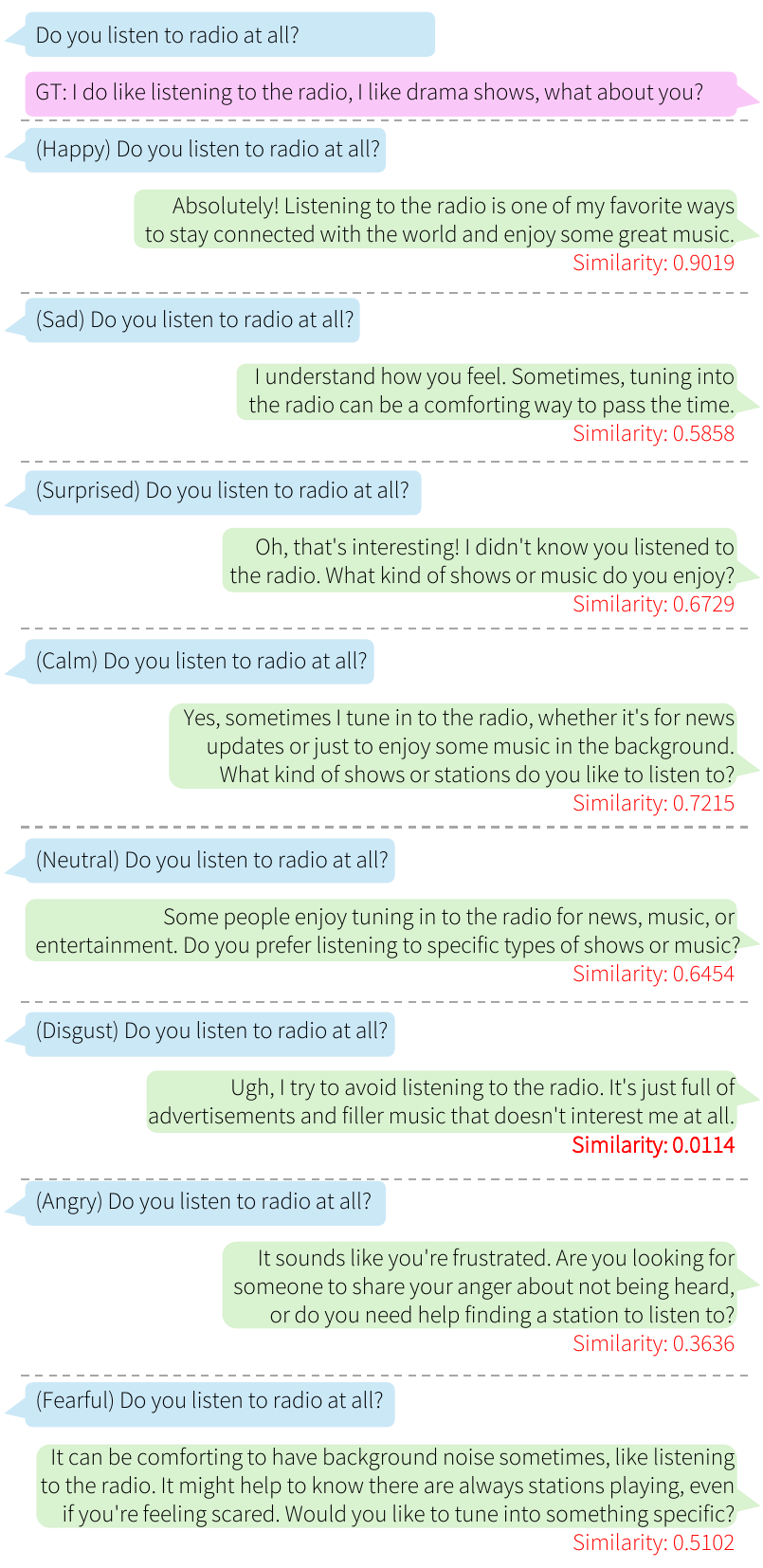}  
    \caption{Examples of Counter-Emotional Response Generation. The ground truth (GT) response is displayed at the top, followed by generated responses for each emotion label. The emotional similarity score, calculated using RoBERTa model trained on GoEmotions dataset, is shown for each generated response.}
    \vspace{-0.2cm}
    \label{fig:counter_emotion1}
\end{figure}

\begin{table}[H]  
\centering
\begin{minipage}{0.95\linewidth}  
\centering
\begin{tcolorbox}
\small
{\color[HTML]{3531FF} \textbf{Counter-Emotional Response \\Generation Instructions:}}\\
Generate a response that align with the emotional tone of a given sentence.
\\
(Happy) Do you listen to radio at all?
\end{tcolorbox}
\vspace{-0.4cm}
\caption{Prompt for the counter-emotional response generation.}
\label{table:prompt_counter_emotional}
\vspace{-0.4cm}
\end{minipage}
\end{table}

In~\cref{fig:counter_emotion1} provides an example of the Counter-Emotional Response task. In this task, we generated responses aligned with the given emotion label. Among the generated responses, we selected the response that is the most emotionally distinct from the ground truth (GT) sentence. This selection process is performed using RoBERTa~\cite{liu2019roberta} model trained on GoEmotions~\cite{demszky2020goemotions} dataset to calculate the emotional similarity between the generated sentences and the ground truth (GT) sentences. The sample with the lowest emotional similarity score, specifically below a threshold of 0.1, is chosen.

As illustrated in~\cref{fig:counter_emotion1}, the responses to the sentence \enquote{Do you listen to radio at all?} vary depending on the specified emotional tone.  For instance, when the emotion label is \enquote{Happy}, the model generates a cheerful and positive response to the sentence \enquote{Do you listen to radio at all?}. It aligns with the positive sentiment of the ground truth, resulting in a high similarity score. Conversely, when the emotion label is \enquote{Disgust}, the model produces a cynical and negative response, significantly differing from the ground truth, leading to a low similarity score. In this example, the response for \enquote{Disgust} label exhibits the lowest similarity score, also falling below 0.1, and is thus selected as the counter-emotional response.

\section{Details of Human Evaluation}
\label{sec:human_evaluation}

\begin{figure}[H]
    \centering
    \includegraphics[width=0.9\columnwidth]{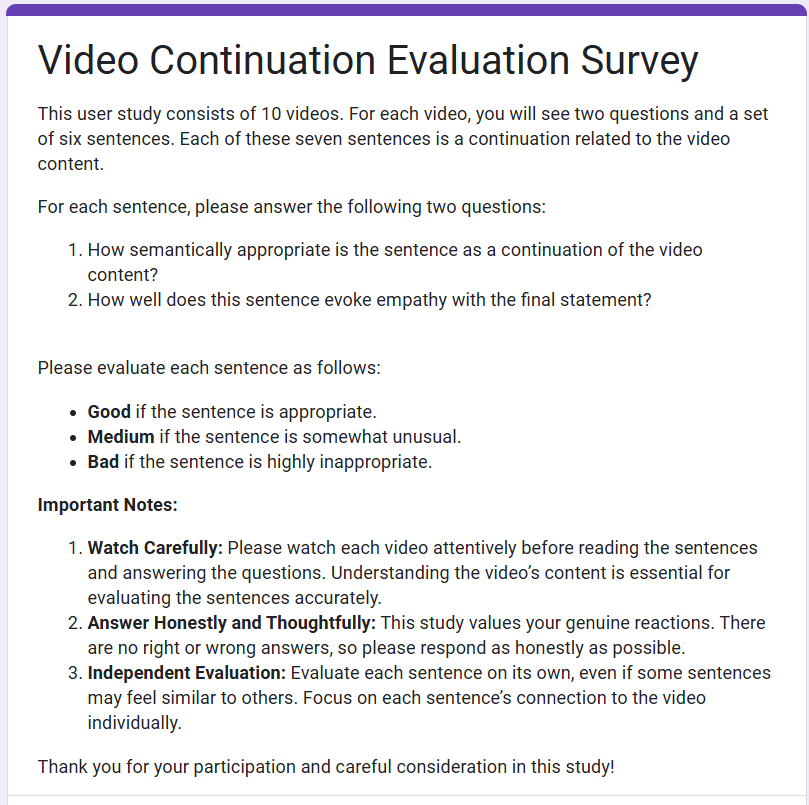}  
    \vspace{-0.2cm}
    \caption{Survey instructions for participants.}
    \vspace{-0.3cm}
    \label{fig:survey1}
\end{figure}

\begin{figure}[H]
    \centering
    \includegraphics[width=0.9\columnwidth]{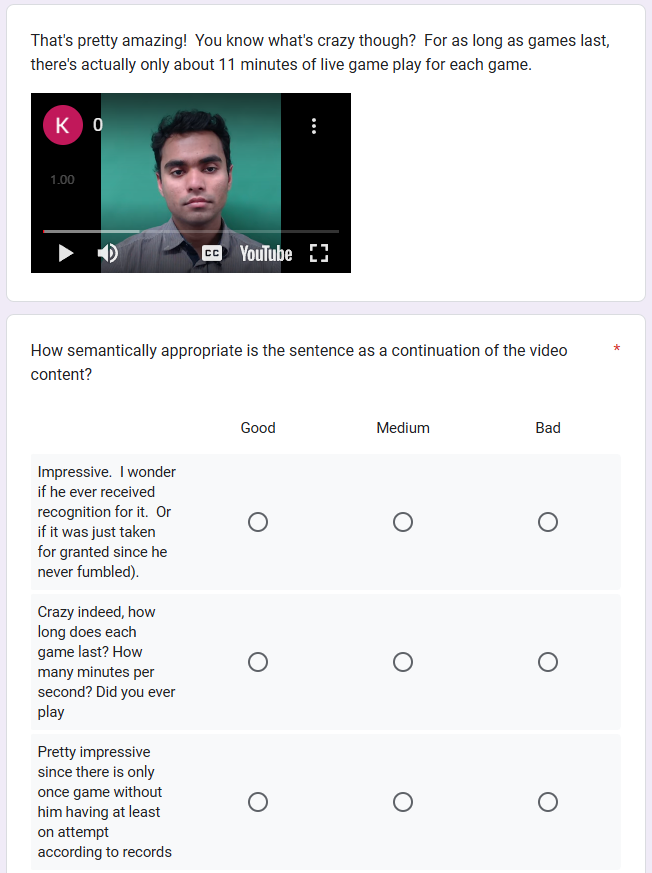}  
    \vspace{-0.2cm}
    \caption{Survey interface for participants. Participants were provided with a conversation video and the responses to the final utterance, generated by six different models.}
    \vspace{-0.3cm}
    \label{fig:survey2}
\end{figure}

We conducted a human evaluation using the Prolific platform. Ten participants were randomly selected to assess ten sample responses, which were selected from the MELD~\cite{poria2018meld}, IEMOCAP~\cite{busso2008iemocap}, and MultiDialog~\cite{park2024let} datasets, generated by our model and comparison models. Each participant was provided with detailed guidelines, as illustrated in~\cref{fig:survey1}. They were presented with a conversation video and its concluding statement, along with responses from six different models, as shown in~\cref{fig:survey2}. Participants evaluated each response for semantic appropriateness and emotional alignment with the preceding context.

\input{tables/table4}

\section{Examples of MambaCompressor's Conversation Restoration}

\begin{figure}[H]
    \centering
    \includegraphics[width=0.85\columnwidth]{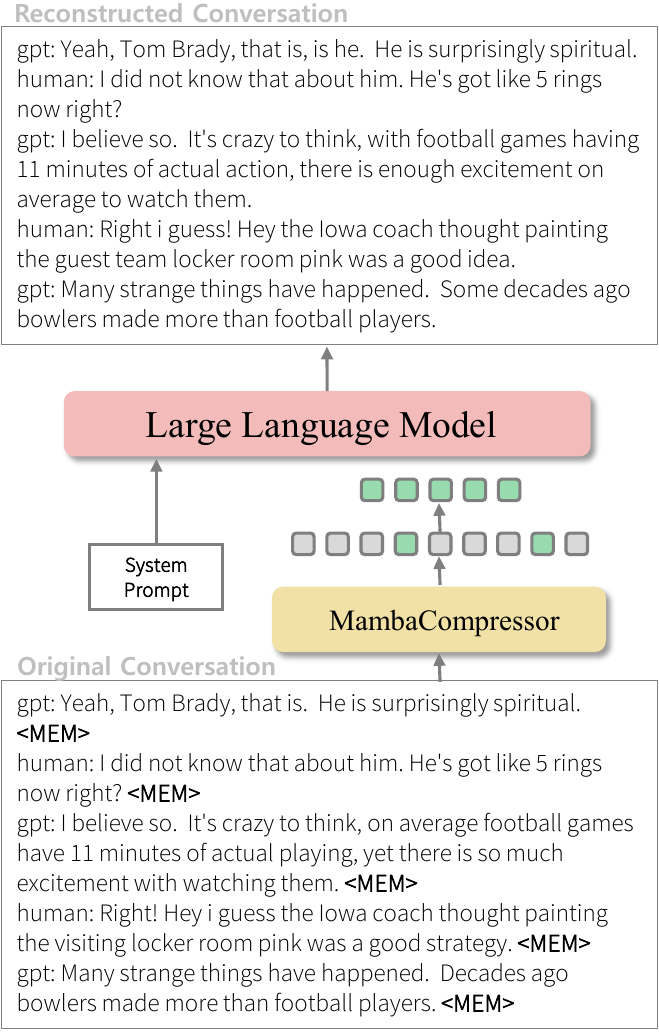}  
    \caption{Example of conversation reconstruction using MambaCompressor.}
    \label{fig:memory1}
\end{figure}

This section presents examples of conversations restored using MambaCompressor. In~\cref{fig:memory1}, we demonstrate how a conversation consisting of five utterances is reconstructed from five MambaCompressor embeddings. By examining the 3rd, 4th, and 5th sentences in~\cref{fig:memory1}, we can observe that although the reconstructed and original sentences are slightly different, their meanings remain the same. This demonstrates that MambaCompressor effectively compresses each utterance while retaining its full content during summarization.

To quantitatively evaluate the conversation reconstruction ability, we measured the BLEU and BERTScore between the original and reconstructed conversations using MultiDialog test-freq split. According to the results in~\cref{table:conversation_reconstruction}, the model achieved a BLEU-1 score of 0.848, a BLEU-4 score of 0.679, and a BERTScore of 0.969, which are considered remarkably high scores. These results indicate the model's strong ability to reconstruct conversations, capturing both n-gram overlap and semantic similarity effectively.

%

%% file: tables/table4.tex
\begin{table}[h!]
\centering
\caption{Automatic evaluation for conversation reconstruction.}
\vspace{-0.2cm}
\resizebox{0.8\linewidth}{!}{
\begin{tabular}{lccc}
\toprule
\textbf{Dataset} & \textbf{BLEU-1} & \textbf{BLEU-4} & \textbf{BERTScore} \\ 
\midrule
MultiDialog & 0.848 & 0.679 & 0.969 \\ 
\bottomrule
\end{tabular}
\label{table:conversation_reconstruction}
}
\end{table}